\documentclass{article}

\usepackage{PRIMEarxiv}

\usepackage[utf8]{inputenc} 
\usepackage[T1]{fontenc}    
\usepackage{hyperref}       
\usepackage{url}            
\usepackage{booktabs}       
\usepackage{amsfonts}       
\usepackage{nicefrac}       
\usepackage{microtype}      
\usepackage{lipsum}
\usepackage{fancyhdr}       
\usepackage{graphicx}       
\graphicspath{{media/}}     

\usepackage{float}
\usepackage{amsmath}
\usepackage{svg}
\usepackage{booktabs}
\usepackage{subcaption}
\usepackage{hyperref}
\usepackage{amsfonts} 
\usepackage{amssymb}

\newtheorem{definition}{Definition}

\title{Tailoring Machine Learning for Process Mining
}

\author{
 Paolo Ceravolo \\
  Computer Science Department  \\
  Universit\`a degli Studi di Milano, Italy  \\
  \texttt{paolo.ceravolo@unimi.it} \\
  \AND
  Sylvio Barbon Junior \\
  Department of Engineering and Architecture  \\
  University of Trieste, Italy \\
  \texttt{sylvio.barbonjunior@units.it} \\
  \And
  Ernesto Damiani \\
  Center for Cyber-Physical Systems  \\
  Khalifa University, Abu Dhabi, UAE\\
  \texttt{ernesto.damiani@ku.ac.ae} \\
  \And
  Wil van der Aalst \\
  Process and Data Science\\
  RWTH Aachen University \\
  \texttt{wvdaalst@pads.rwth-aachen.de} \\
}

\begin{document}
\maketitle

\begin{abstract}
Machine learning models are routinely integrated into \textit{process mining} pipelines to carry out tasks like data transformation, noise reduction, anomaly detection, classification, and prediction. Often, the design of such models is based on some \emph{ad-hoc} assumptions about the corresponding data distributions, which are not necessarily in accordance with the \textit{non-parametric} distributions typically observed with process data. Moreover, the learning procedure they follow ignores the constraints \textit{concurrency} imposes to process data. Data \textit{encoding} is a key element to smooth the mismatch between these assumptions but its potential is poorly exploited. In this paper, we argue that a deeper insight into the issues raised by training machine learning models with process data is crucial to ground a sound integration of process mining and machine learning. Our analysis of such issues is aimed at laying the foundation for a methodology aimed at correctly aligning machine learning with process mining requirements and stimulating the research to elaborate in this direction.
\end{abstract}


\section{Introduction}
\label{intro}

Process Mining (PM) is a consolidated discipline grounded on \textit{data mining} and \textit{business process management}. The exploitation of traditional PM tasks (\textit{discovery}, \textit{conformance checking}, and \textit{enhancement}) is today a reality in many organizations~\cite{Deloitte2021,van2011process}. In the last decade, a wave of new results in \textit{artificial intelligence} has triggered the interest of the PM research community in using supervised or unsupervised Machine Learning (ML) techniques for gaining insight into business processes and providing advice on how to improve their inefficiencies. 

In today's practice, ML models are routinely integrated into PM data pipelines~\cite{van2015processes} to carry out tasks like data transformation, noise reduction, anomaly detection, classification, and prediction. For example, ML is playing a key role in the interface between PM and sensor platforms. Advances in sensing technologies have made it possible to deploy distributed monitoring platforms capable of detecting fine-grained events. The granularity gap between these events and the activities considered by classic PM analysis has often been bridged using ML models \cite{tax2016event,van2021event} that compute virtual activity logs, a problem which is also known as \textit{log lifting}~\cite{tello2019machine}. ML has been proposed as a key technology to \textit{strengthen} existing techniques, for example, using trace clustering to reduce the diversity that a process discovery algorithm must handle in analyzing an event log~\cite{song2008trace,bose2009context,appice2015co,tavares2022selecting}, to simplify the discovered models ~\cite{kalenkova2020framework,senderovich2018aggregate,chapela2019simplification}, or to support real-time analysis on event streams~\cite{mishra2019analysis,tavares2019overlapping,ceravolo2020evaluation}. ML is adopted to apply predictive models to the executing cases of a process. This research area, known as \textit{predictive process monitoring}, exploits event log data to foresee future events, remaining time, or the outcome of cases, in support of decision making~\cite{10.1007/978-3-030-35166-3_25,pasquadibisceglie2020predictive,marquez2017predictive}. Root cause analysis~\cite{bozorgi2020process} and data explainability~\cite{hanga2020graph} are other tasks that can be applied to event log data using  ML techniques, in order to improve our understanding of a business process.
ML models have also been used in addition to (or in lieu of) classic linear programming \cite{Wiel2010ProcessMU} to \textit{optimize} business processes' resource consumption and to provide insights to process \textit{re-design}~\cite{al2021using}.
Computational support for PM tends to converge with the one available for ML models also from the technology standpoint~\cite{van2020academic,veit2017proactive}. This makes their integration seem straightforward. 

In fact, it is not. When PM tasks are mapped to ML tasks, business process-specific assumptions should drive the construction of training functions and hyperparameters selection. Some of these assumptions stem from the very nature of human social systems. For example, it is well known that process variants are shaped by \textit{non-parametric} distributions~\cite{van2020pareto}. Quite the contrary, data normality is beneficial for many ML models, moreover, if the data distribution is skewed, ML models may be biased toward a particular outcome. In addition, the ML view on event log data is often oversimplified. The correct encoding of the procedural nature of event log traces is challenging. Often, the sequence of executed events is simply captured by a prefix of fixed length. Even more problematic is encoding \textit{concurrency} and the \textit{interactions} constraining the events in the business process. Encoding event log data into a feature space compatible with ML algorithms is a critical design choice in other concerns~\cite{barbon2020evaluating}. It impacts the \textit{sample complexity}, the \textit{data distribution}, and the relevance of the features to put under analysis, for example, to detect \textit{concept drift} or to support \textit{zero-shot learning}~\cite{hilprecht2021one}.

Today, much of the research on integrating ML with PM focuses on developing ML models to attain high performance in specific business process management scenarios. Less attention has been paid to designing a general methodology to select and adapt ML models based on the nature of the PM problem, taking into account the specific properties of the process data. We argue that, when using ML models in PM pipelines, it is important to prevent any \textit{mismatch} between the assumptions on input data distributions underlying the ML models and the statistics of the event logs used to feed them~\cite{marques2023meta}. Arbitrarily selecting algorithms leads to unfair evaluation and sub-optimal solutions. For example, a given model cannot be compared with another if their implementations consider different feature spaces~\cite{ManeiroVL21}. It is also important to make sure that ML models are exposed to process-specific information, such as the processes' control-flow constraints. In this paper, we attempt to identify some of the causes of this mismatch and suggest how to remove them, with the aim of fostering research on a sound methodology to address the integration between PM and ML.

We believe that an effort on these aspects must be jointly made by the PM and Artificial Intelligence research communities. This call to collaboration is valid in general but particularly in business process management, where data analysis has to leave the safe harbor of experimental science to sail into the open sea of decision science. 
In this paper, we discuss the challenges in a specific direction, i.e., from PM to ML.
More specifically, in Section \ref{landscape} we discuss the issues leading to the PM-to-ML mismatch. In Section \ref{basic} we introduce some basic PM notions. In Section \ref{bridge} we link them to ML principles. Section \ref{illustrative} clarifies the discussion by presenting a couple of samples. Section \ref{method} proposes research lines for advancing in the direction of a general methodology that integrates ML models into PM pipelines. Section \ref{conc} closes the paper.

\section{The Issues Landscape}
\label{landscape}

An important problem underlying our discussion is how to take into account process data specificity in ML model selection and (hyper-) parameter tuning. Of course, processing event logs poses all the usual challenges of data pre-processing and preparation. We will not discuss standard data pre-processing techniques such as outlier removal~\cite{sani2018applying,sani2019repairing}, noise filtering~\cite{li2018anti,sun2019filtering}, and missing entries recovery~\cite{fox2018data} as they can be tackled by current statistical techniques.
Rather, we will focus on issues that are specific to process data, including their statistical distribution and event concurrency. Indeed, careless assumptions on the encoding of input data may result in biased models with reduced generalization capability. 

\subsection{Data Distribution}
When choosing an ML model for a PM task, it is tempting to assume that the process data fed to the model will follow a normal distribution. Indeed, data normality is beneficial for many types of ML models.
Models like Gaussian, naive Bayes, logistic and linear regression explicitly rely on the assumption that the data distribution is bi-variate or multivariate normal. Many phenomena of interest for business process analysis, such as the duration of some activities, are known to follow normal or log-normal distributions \footnote{See, for instance, the ``lunch break'' duration distributions at \url{https://www.statista.com/statistics/995991/distribution-of-lunch-breaks-by-length-in-europe/}}. 
For other PM data, however, assuming normality is not always advisable. For example, process variants are specific activity sequences that occur through a process from start to end. Variants' occurrence in an activity log is typically following a \textit{non-parametric} trend that complies with the \textit{Pareto principle}~\cite{van2020pareto}. 
A normal distribution cannot always be assured also for the pairwise dependency relationship between activities, a key statistical information exploited by process discovery algorithms~\cite{berti2017statistical}.  
Indeed, in this case, the normality assumption has been verified for some event logs, including some popular benchmarks we will discuss in Section \ref{illustrative} (the ``Road traffic fines''~\cite{deroad} and ``Receipt phase of an environmental permit application process''~\cite{Burece}). However, the normality of dependencies in less regular, ``spaghetti'' like, processes is not observed, as in the ``BPI Challenge 2015 Municipality 1''\cite{Domun}. 
There are reasons to believe that dependencies in loosely specified logs may follow some power-law trend as well, and require careful parameter fitting in statistical analysis.  Imbalanced data sets or non-stationary environments may also cause serious difficulties. For example, if the training data is skewed towards a particular class or outcome, the model may be more likely to predict that class or outcome even when it is not the most likely one. Independent component analysis \cite{lee1998independent} provides ways to reveal Gaussianity and non-Gaussianity. Of course, non-normal distributions can be transformed to normal ones using Box-Cox transformations~\cite{sakia1992box}, and unbalanced data sets can be balanced ~\cite{bifet2015efficient,roccetti2021alternative} but, as we shall see, such data transformations should be applied with caution, as they have consequences on the performance of the models. 

In any case, PM data regarding distributions of variants cannot be expected to always follow a Gaussian behavior, demanding estimation techniques to sample from the sequential process underlying log generation.
\emph{Markov Chain Monte Carlo }(MCMC) techniques are sometimes used for sampling from an unknown probability distribution (for instance, the distribution of variants) by using data to construct a Markov chain whose equilibrium distribution approximates the unknown one. MCMC techniques can be combined with Kalman filtering\cite{emerick2011combining} to control uncertainty.
Of course, an explicit estimate of the data distribution may not even be necessary. Some ML models work well also in the case of non-normally distributed data.  Simple yet effective ML models like decision trees and random forests do not assume any normality and work reasonably well on raw event data. Also, linear regression is statistically effective if the model errors are Gaussian, an assumption less stringent for process data than the normality of the entire data set. Kernel methods, e.g., Gaussian processes and support vector machines, provide flexible models that are practical to work with but require proper hyperparameter variables to fit the data. 


\subsection{Concurrency}
Another key attention point is concurrency. How to use ML to predict the behavior of highly concurrent systems and processes is still an open problem, and the research done in the AI community has only scratched its surface (see \cite{neu2021systematic} for a recent review). Most ML approaches view event logs as merely sequential data~\cite{van2021concurrency}, rather than sequential manifestations of a concurrent system. This may lead to under-sampling the log space and to insufficient training to handle apparently out-of-order event sequences~\cite{di2017eye}. 
To address this issue, it is important to provide ML models with control-flow information about the iterative or concurrent execution of tasks as additional context alongside event logs. One approach that has been explored is the use of Bi-directional Long-Short Term Memory (BiLSTM) architectures. Thapa et al. \cite{thapa2020deep} leveraged BiLSTM to detect concurrent human activities in a smart home environment. Additionally, Thapa et al. \cite{thapa2021adapted} adapted the LSTM algorithm into a synchronous algorithm called sync-LSTM, enabling the model to handle multiple parallel input sequences and generate multiple synchronized output sequences.
The field of predicting the behavior of highly concurrent systems using ML is rapidly evolving, as indicated by the recent survey conducted by Neu et al. \cite{neu2022systematic}. Researchers are actively exploring new techniques and methodologies to improve the understanding and prediction of concurrency in various domains.


\subsection{Non-stationary Behaviour}
Even when the process data distributions can be fitted precisely, running processes, especially the ones involving resources that learn and age like people and equipment, change over time. This gives rise to \textit{non-stationary} behavior. 
This problem is a critical one since ML models' learning capacity decreases under non-stationary conditions~\cite{ceravolo2020evaluation}. Concept drift detection techniques are therefore required. In traditional data mining applications, \textit{concept drift} is identified when, at two separate points in time, a concept, i.e., the relation between a data instance and its associated class, changes~\cite{krawczyk2017ensemble}. In PM, many aspects of drift should be carefully monitored, including the appropriateness of the event trace with respect to the model, the dependency relationship between activities, and the interdependence between the activities and the available resources or cycle time. Each aspect should be appropriately encoded and monitored using statistical analysis ~\cite{baier2020handling}.  

\subsection{Zero-shot Learning}
A related topic is using ML models to identify solutions never observed during training, the so-called \textit{zero-shot learning}~\cite{kappel2021leveraging}. There are several zero-shot learning approaches, but a commonality is that unstructured auxiliary information is encoded during the training process instead of using explicit labels. The training process aims to learn to connect new input elements to encodings that have the greatest similarity in terms of auxiliary information.
In this way, the system can propose an outcome never observed during the training stage.  Zero-shot is relevant in PM when the availability of labeled process data is limited, as the process may be recently developed, unused, or its outcomes inaccessible. In these situations, relying on historical observations to guide learning tasks is insufficient or erroneous. This scarcity has drawn the attention of the PM community to \emph{contrastive learning}, a manner of unsupervised learning that learns representations by
contrasting positive and negative pairs. Graph-related contrastive
learning methods apply this notion to all types of graph data. Some popular unsupervised representation learning
methods imply the idea of contrastive learning. For instance, DeepWalk \cite{perozzi2014deepwalk} and node2vec \cite{grover2016node2vec} generate Markov chains of nodes based on random walking on graphs,
forcing the neighboring nodes of a graph to have similar representations. More recent proposals such as DGI
\cite{velickovic2019deep} and InfoGraph \cite{sun2019infograph} combine contrastive learning with ordinary supervised training to maximize the mutual information of node and graph levels.

Much work is also being done on \textit{generative} engines for logs based on likelihood-based models, like auto-encoders and Generative Adversarial Networks (GANs)\cite{krajsic2021variational}.   However, ordinary GANs show some 
limitations when applied to generate ``clean'' process data, where low confidence variants are due to failures of the monitoring context rather than to adversarial constructions \cite{song2019generative}. In addition, the GANs' \emph{objective function}, i.e. the difference between the generated and the original distribution of traces in the event log, is not always suitable for evaluating the quality of the generated process variants, and even less for comparing different generators. Performance measures should be used instead, and the trained algorithms should be able to provide an answer with different information details, for example, predicting a performance result knowing or not the availability of resources currently in use.


\subsection{Data Encoding}
Supervised ML algorithms are trained on collections of examples, each encoded as a vector in a multidimensional feature space. An appropriate encoding method can reduce the sample complexity and reduce the space or time complexity of the model~\cite{barbon2020evaluating}.
In PM, even more, than selecting individual features, it is important to capture the interconnections between the different process dimensions. 
The event logs analyzed in PM contain information from several complementary dimensions, such as event data, executing traces, resource consumption, and cycle time. Each event can be described as a multidimensional object, but its value for the process execution lies in the interdependence with the other events composing the process case instance, the resources available in the system, and the temporal limits constraining the case, which in turn depend on the other cases executed, executing, or to be executed in the system. Therefore, capturing the constraints due to the alternative, optional or mandatory dependency between events is crucial in PM. 
Encoding methods should also identify features subject to \textit{concept drift}. 
Extracting insights from this type of functional data is not straightforward; covariance control\cite{wang2022low} is needed to take into account the hidden relationships between the different dimensions.

Despite all this, little effort has been spent by the PM community to study the impact of encoding methods on the performance of PM pipelines. Only a few comparative studies are available~\cite{barbon2023trace,barbon2020evaluating,teinemaa2019outcome,KoninckBW18}. 
Basic techniques, such as the one-hot encoding scheme~\cite{TaxVRD17} or frequency-based encoding ~\cite{appice2015co}, are often adopted. For numerical attributes, general statistics have been used, such as average, maximum, minimum, and sum~\cite{pasquadibisceglie2020predictive}. 
The $k$-gram encoding schema~\cite{bose2009context} is also quite popular. Each activity in the trace is represented as the sequence of $k$ activities executed to reach it. As an alternative, arrays encoding traces as the frequency of their activities at each position have been proposed~\cite{ceravolo2017toward}. These encoding techniques can incorporate some control-flow information, but cannot fully account for concurrency. 
To better capture dependency between activities,  techniques borrowed from other domains have been proposed, including text mining~\cite{WeissIZ15,Le2014} and graph embedding~\cite{Grover2016,Perozzi2017}.
Graph embedding methods emerged from the necessity of representing graphs as low-dimensional vectors to be exploited by downstream ML models. These methods rely themselves on ML models (usually, supervised learners) to compute highly informative but low-dimensional vectors of fixed length~\cite{Azzini2021}. When applied to event logs encoding, such methods outperform the others, at the cost of higher time complexity and loss of transparency, as the resulting vectors are organized in a latent space losing any reference to the event log attributes or their statistical properties~\cite{tavares2020analysis}. In any case, the representation of the control-flow is purely sequential and concurrency is not captured by these methods too. 
Recently, emerging attention on techniques for encoding control-flow information into a feature space is observed, for example by representing the degree of parallelism or optionality of activities ~\cite{ChiorriniDGPP22,vazifehdoostirani2022encoding}. Another trend is aimed at constructing multi-perspective views of traces, representing the data-flow and control-flow into the same encoding~\cite{pasquadibisceglie2021multi,camargo2019learning}.  However, the application of these methods is still limited. 

Generally speaking, the encoding procedures used to map PM data to ML models are not documented enough in the PM literature. Sometimes, the feature space selected is not explicitly presented, the steps followed to encode data are not well specified, or the adopted code is not shared. \emph{Ablation studies}, removing parts of the data representation and studying the removal's impact on performance, are still the exception rather than the norm.
We argue that the formalization of the encoding procedure allows explaining this key design choice to be justified by the specific analytical goals and the assumptions applying to the algorithms considered.  We will propose such a formalization in Section \ref{bridge}.

\section{Basic Notions in PM}
\label{basic}

To make this paper self-contained, in this section we recall some of the basic concepts of PM. 
An {\em event log} is a collection of {\em events} generated in a temporal sequence and stored as {\em tuples}, i.e., recorded values from a set of {\em attributes}. Events are aggregated by {\em case}, i.e., the end-to-end execution of a business process. For the sake of classification, all cases following the same {\em trace}, i.e., performing the same sequence of business process activities, can be considered equal as they belong to the same process \textit{variant}. 

\begin{definition}[Event, Attribute]\label{def:event}
Let $\Sigma$ be the \textit{event universe}, i.e., the set of all possible event identifiers; $\Sigma^{*}$ denotes the set of all finite sequences over $\Sigma$. Events have various \textit{attributes}, such as \textsc{timestamp}, \textsc{activity}, \textsc{resource}, \textsc{associated cost}, and others. Let $\mathcal{AN}$ be the set of attribute names. For any event $\textit{e} \in \Sigma$ and  attribute $\textsc{a} \in \mathcal{AN}$, the function $\#_\textsc{a}(\textit{e})$ returns the value of the attribute \textsc{a} for event \textit{e}. 
\end{definition}

The set of possible values of each attribute is restricted to a domain. For example, $\#_\textsc{activity}: \Sigma \rightarrow \mathcal{A}$, where $\mathcal{A}$ is the set of the legal activities of a business process, e.g. $\mathcal{A} = \{a,b,c,d,e\}$. 
If $\textit{e}$ does not contain the attribute value for some $\textsc{a} \in  \mathcal{AN}$, then $\#_{\textsc{a}}(\textit{e})= \bot$.
It follows that an event can also be viewed as a tuple of attribute-value pairs $e = (\mathcal{A}_1, ..., \mathcal{A}_m)$, where $m$ is the cardinality of $\mathcal{AN}$.

\begin{definition}[Sequence, Sub-sequence]\label{def:trace}
In a sequence of events $\sigma \in \Sigma^\ast$, each event appears only once and time is non-decreasing, i.e., for $1 \leq \textit{i} \leq \textit{j} \leq |\sigma|: \#_\textsc{timestamp}(e_{i}) \leq \#_\textsc{timestamp}(e_{j})$. 
Thus $\langle e_{1},e_{2},e_{3} \rangle$ denotes three subsequent events. 
A sequence can also be denoted as a function generating the corresponding event for each position in the sequence: $\sigma(i \rightarrow n) \mapsto \langle e_{i}, ..., e_{n}\rangle$, with $e_{n}$ the last event of a sequence. In this way, we can define a sub-sequence as a sequence $\sigma(i \rightarrow j)$ where $0 \leq i <j<n$.
\end{definition}

\begin{definition}[Case, Event Log]\label{def:case} 
Let $\mathcal{C}$ be the \textit{case universe}, that is, the set of all possible identifiers of a business case execution. $\mathcal{C}$ is the domain of an attribute $\#_\textsc{case}  \in \mathcal{AN}$. We denote a case $c \in \mathcal{C}$ as $\langle e_{1},e_{2},e_{3} \rangle_{c}$, meaning that all events are in a sequence and share the same case.  For a case $\langle e_{1},e_{2},e_{3} \rangle_{c}$ we have $\#_\textsc{case}$(\textit{$e_{1}$}) = $\#_\textsc{case}$(\textit{$e_{2}$}) = $\#_\textsc{case}$(\textit{$e_{3}$}) = $c$. An \textit{event log} $L$ is a set of cases $L \subseteq \Sigma^{*}$ where each event appears only once in the log, i.e., for any two different cases, the intersection of their events is empty. When the case identifier is not used as a grouping attribute, an \textit{event log} $\hat{L}$ can be simply viewed as a set of events, thus $\hat{L} \subseteq \Sigma$.
\end{definition}

\begin{definition}[Variant, Event Log]\label{def:var}
The cases $c_1$ and $c_2$ follow the same variant if $\langle e_{1},e_{2},e_{3} \rangle_{c_1}$ and $\langle e_{4},e_{5},e_{6} \rangle_{c_2}$ have the same sequence of activities, e.g.  $\#_\textsc{activity}$(\textit{$e_{1}$}) = $\#_\textsc{activity}$(\textit{$e_{4}$}) = $a$, $\#_\textsc{activity}$(\textit{$e_{2}$}) = $\#_\textsc{activity}$(\textit{$e_{5}$}) = $b$, $\#_\textsc{activity}$(\textit{$e_{3}$}) = $\#_\textsc{activity}$(\textit{$e_{6}$}) = $a$. We call this sequence a trace. This implies an event log can also be viewed as a multi-set of traces. We denote an event log as a multi-set by writing $\overline{L} = [\langle a,b,c  \rangle^{3}, \langle a,b,a  \rangle^{11}, \langle a,c,b,a \rangle^{20}]$. The superscript number of a trace details the number of cases following this variant. For example,  $\langle a,b,a \rangle^{11}$ means we have a variant with $11$ cases following the trace $\langle a,b,a\rangle$. 
\end{definition}

\section{A Formalisation of  PM Data Encoding}
\label{bridge}

Despite the variety of encoding methods discussed in Section~\ref{landscape}, we argue that available approaches fail to capture key process-level information such as the interplay between cases, or between activity execution and availability of resources. Most of the encoding methods  in use today focus on the \textit{control-flow}, according to an \textit{inter-case} view. Methods focusing on the \textit{intra-case} view have been proposed but are rarely applied ~\cite{SenderovichFGJM17}. Similarly, proposals for encoding the \textit{data-flow}~\cite{LeoniA13} are available in the literature, but never adopted in comparative studies or surveys. Another recent trend is stressing the need of capturing constraints connected to concurrency~\cite{ChiorriniDGPP22,vazifehdoostirani2022encoding}.
In this section, we discuss in detail how PM data is encoded to suit ML models' training procedures. 
For the sake of space, we limit our discussion to supervised learning, probably the most widely applied ML approach.
Generally speaking, supervised techniques train models to compute functions $f:  \mathbb{R}^d \rightarrow \mathbb{R}^{d'}$ where the input is a $d$-dimensional vector $\mathbf{x}$ and the output is a $d'$-dimensional vector $\mathbf{y}$. Each dimension is a measurable piece of data, a.k.a feature or attribute. For popular ML tasks, the output is mono-dimensional. In regression,  the output is a real-valued scalar value, while in classification,  the output is a natural number indexing a ``class''. However, nothing prevents having multidimensional vectors in output. In structured learning, input and output may be a structure like a block matrix, divided into sub-matrices to represent algebraic entities such as graphs, tensors, etc.
The training process to approximate $f$ requires a set of examples $\{(\mathbf{x}_1, \mathbf{y}_1), ..., (\mathbf{x}_n, \mathbf{y}_n)\}$ where inputs and outputs are paired. We can then define this training set as an example matrix $\mathbf{X} := [\mathbf{x}_1, ..., \mathbf{x}_n]^{\top} \in \mathbb{R}^{n \times d}$ and a label matrix $\mathbf{Y} := [\mathbf{y}_1, ..., \mathbf{y}_n]^{\top} \in \mathbb{R}^{n \times d'}$, given by the number $n$ of vectors and the number $d$ of dimensions in the vector space.

 In their original format, PM log entries do not belong to a vector space. This is because the events in an event log are grouped by case and this grouping is essential to keep a connection with business process execution.

Our goal here is to formalize the procedure to encode the cases into vectors in a way that can be used as a template to describe the specific encoding chosen for a PM application. Our starting point is $\hat{L} \subseteq \Sigma$, a log view as a set of event identifiers. This representation can be mapped into a vector space $\mathbf{X}$ by applying a suitable \emph{transformation function} grouping event by case and returning vectors of size equal to or less than the event size. 

\begin{definition}[Encoding  function]\label{def:project}  
Given an event log $\hat{L} \subseteq \Sigma$, an \emph{encoding function}  $\Gamma: \Sigma \rightarrow \mathbf{X}^{n \times d}$ represents $\hat{L}$ in the vector space $\mathbf{X}$.
The encoding function $\Gamma$ is valid if it defines a transformation where two elements of $\Sigma$, $e_i$ and $e_j$ are aggregated on the same element $\mathbf{x} \in \mathbb{R}^d$ if $\#_\textsc{case}$(\textit{$e_{i}$}) = $\#_\textsc{case}$(\textit{$e_{j}$}), with $n \leq |\mathcal{C}|$, i.e. the vectors in $\mathbf{X}$ are a subset of the cases in  $\mathcal{C}$.
\end{definition}

We propose a canonical representation of $\Gamma$ as a composition of a \emph{filtering function} $\pi$, a \emph{dimensioning function} $\rho$, a \emph{grouping function} $\eta$, and a \emph{valuation function} $\nu$, i.e., $\Gamma= \nu \circ \eta \circ \rho \circ \pi$.
One or more of these components can implement the identity function with null effects.

In particular, $\pi: \Sigma \rightarrow \Sigma_{\alpha}$ imposes a condition on the events' attributes or the attributes' values, $\forall e \in \hat{L} \land \textsc{a} \in  \mathcal{AN} : P(\#_{\textsc{a}}(e)) $, where $P$ is a predicate, thus $|\Sigma_{\alpha}| \leq |\Sigma|$. For example, filtering the events by their timestamp $\forall e \in \hat{L} :$ {\tt YYYY-MM-DD} $\geq \#_\textsc{timestamp}(e) \leq$ {\tt YYYY-MM-DD}. The function $\rho: \Sigma_{\alpha} \rightarrow D$ defines the dimensions of the vector space, creating new dimensions based on a range of values in the original dimensions or, less commonly,  grouping multiple dimensions into a single one. Often, the set $D$ is the union of multiple attribute domains, i.e. $D= \mathcal{A}_{k=1}  \cup \mathcal{A}_{k=2} \cup \cdots \mathcal{A}_{k=l}$. 
The function $\eta: \Sigma_{\alpha} \rightarrow \mathbf{X}^{n \times d}_{\alpha}$, with $d = |D|$, assigns to $\mathbf{X}_{\alpha}$ the values of the attributes in $e$ and groups events by case so that $\forall \mathbf{x} \forall \textsc{a}_k : \mathbf{x}_{i,j} = \#_{\textsc{a}_k}(\textit{e}) \iff  \#_{\textsc{a}_k}(\textit{e}) = D_j \land \#_\textsc{case}(e) = c_i$. The number of elements in the vector space equals the number of cases to include in the example matrix, thus $n \leq |\mathcal{C}|$. Because the sets $\Sigma_{\alpha}$ and $D$ can be view as columnar matrices $M_{\Sigma_{\alpha}}^{n \times 1}$ and $M_{D}^{d \times 1}$, the size of $\mathbf{X}_{\alpha}$ is equal to $M_{\Sigma_{\alpha}} \times M_{D}^{\top}$, i.e. the set of events we selected with $\pi$ is multiplied by the dimensions we identified with $\rho$. It is worth mentioning that, when grouping is applied, each vector component becomes an array of attribute values rather than a single value. The function $\nu$ aims at transforming these arrays of attribute values into real-valued scalar values. We define $\nu: \mathbf{X}^{n \times d}_{\alpha} \rightarrow \mathbf{X}^{n \times d}$ to clarify the components of the two matrices are valuated differently.




For example, the basic \textit{one-hot} encoding schema corresponds to a null $\pi$, a $\rho$ with $D = \bigcup_{k=1}^{l} \mathcal{A}_{k}$, an $\eta$ for grouping the events of a same case, and a
$\nu: \mathbf{X}^{n \times d}_{\alpha} \rightarrow \{0,1\}^{n \times d}$, returning $\mathbf{x}_{i,j} = 1$ if at least a value $\#_{\textsc{a}_k}(\textit{e}) = D_j$ is observed for the case $\#_\textsc{case}(e) = c_i$, and $0$ if not. 
The popular \emph{activity profile} schema~\cite{song2008trace} encodes an event log into a vector of activity values by simply counting all events of a case that include that activity. The encoding function maps the events in $\hat{L}$ into $\mathbf{X}$ by executing the four canonical transformations as follows.
First, it verifies to consider only events associated with activity values  $\forall e \in \Sigma : \#_\textsc{activity}(\textit{e}) \neq \bot$. Then it defines the dimensions of $\mathbf{X}$ with $\rho$ so that $D = \mathcal{A}$, where $\mathcal{A}$ is the set of legal business process activities. Third, it aggregates the data by case with $\eta$. Finally, it performs the evaluation with $\nu$, assigning the count of the components in $\mathbf{x}_{i,j}$ for each case $c_i$. For instance, the log $\overline{L} = [\langle a,b,c  \rangle^{3}, \langle a,b,a  \rangle^{11}, \langle a,c,b,a \rangle^{20}]$, is transformed in the first matrix in \ref{m1} with $\pi$, in the second matrix with $\rho$, in the third matrix in with $\eta$, to finally get the fourth matrix in \ref{m1} with $\nu$.

\begin{equation}
\begin{bmatrix}
e_1 \\
e_2 \\
e_3 \\
e_4 \\
e_5 \\
...
\end{bmatrix}
\begin{bmatrix}
a\\
b\\
c\\
\end{bmatrix}
\begin{bmatrix}
a & b & c\\
a & b & c\\
a & b & c\\
[a,a] & b & \bot\\
[a,a] & b & \bot\\
...
\end{bmatrix}
\begin{bmatrix}
1 & 1 & 1\\
1 & 1 & 1\\
1 & 1 & 1\\
2 & 1 & 0\\
2 & 1 & 0\\
...
\end{bmatrix}
\label{m1}
\end{equation}

We believe that if the PM community would get used to clarifying the definition of the following functions when defining an encoding procedure, the literature will benefit in terms of the comparability of the results. For example, a \textit{data-fow} approach will require clarifying the contribution of the different dimensions in encoding cases. An \textit{intra-case} approach will require modifying the $\eta$ function to encode multiple cases into a single vector.

\begin{table}[]
\centering
\begin{tabular}{@{}lll@{}}
\toprule
Cases & Number of Variants & Coverage of Cases \\ \midrule
56482 & 1 & 37,6\%  \\
102853	& 2	& 68,4\% \\
132758	& 4	& 88,3\% \\
142926	& 7	& 95,0\% \\
148887	& 17 &	99,0\% \\
150270	& 131 &	99,9\% \\
150370	& 231 &	100,0\% \\
  \\ \bottomrule
\end{tabular}
  \caption{\textit{Managing Road Traffic Fines} Event Log}
  \label{tab:1}
\end{table}

%
%
%
%
%

\begin{figure}
\begin{subfigure}{.49\textwidth}
  \centering
  \includegraphics[width=.95\linewidth]{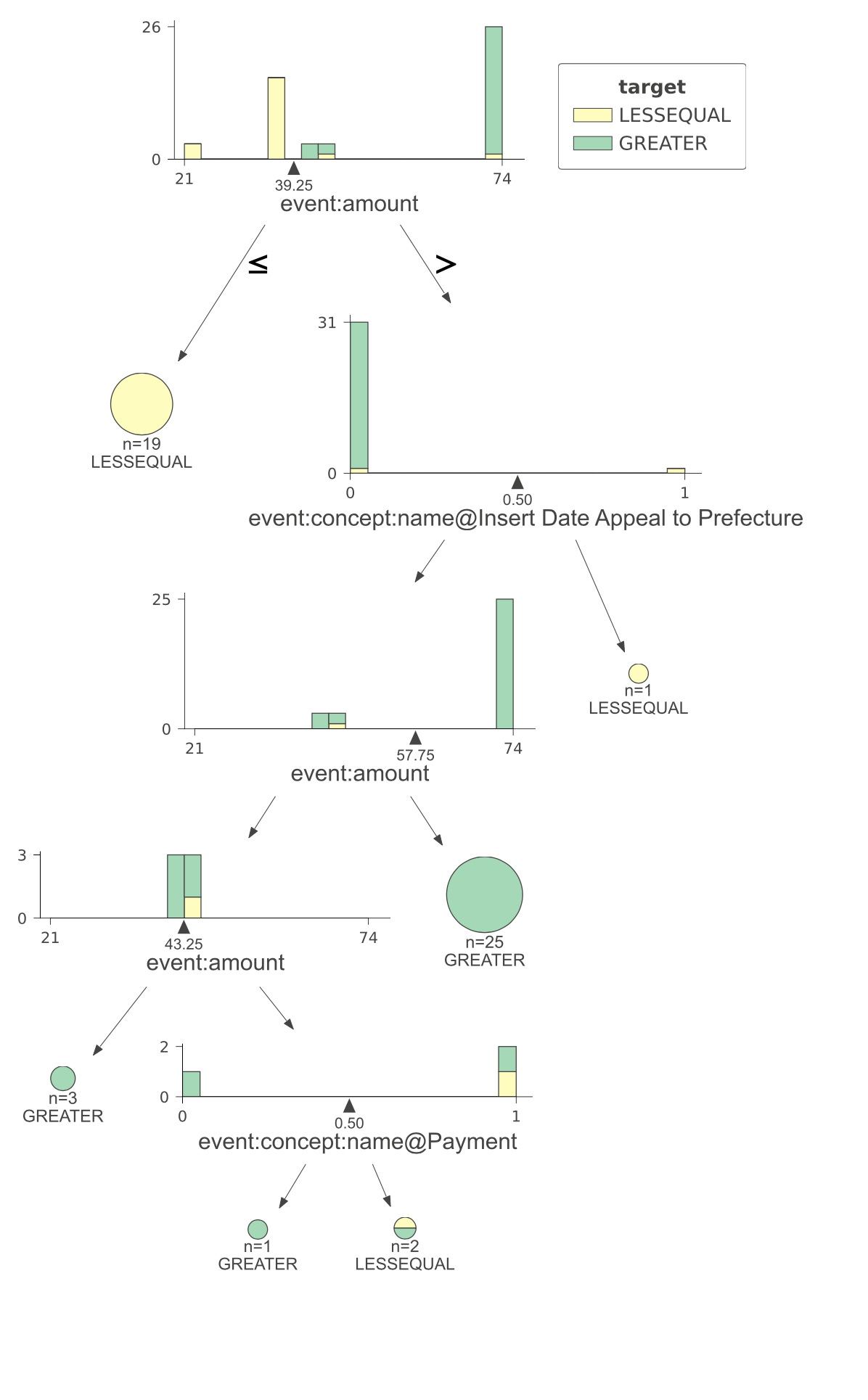}
  \caption{Unbalanced event log}
  \label{fig:DTsfig1}
\end{subfigure}%
\begin{subfigure}{.49\textwidth}
  \centering
  \includegraphics[width=.95\linewidth]{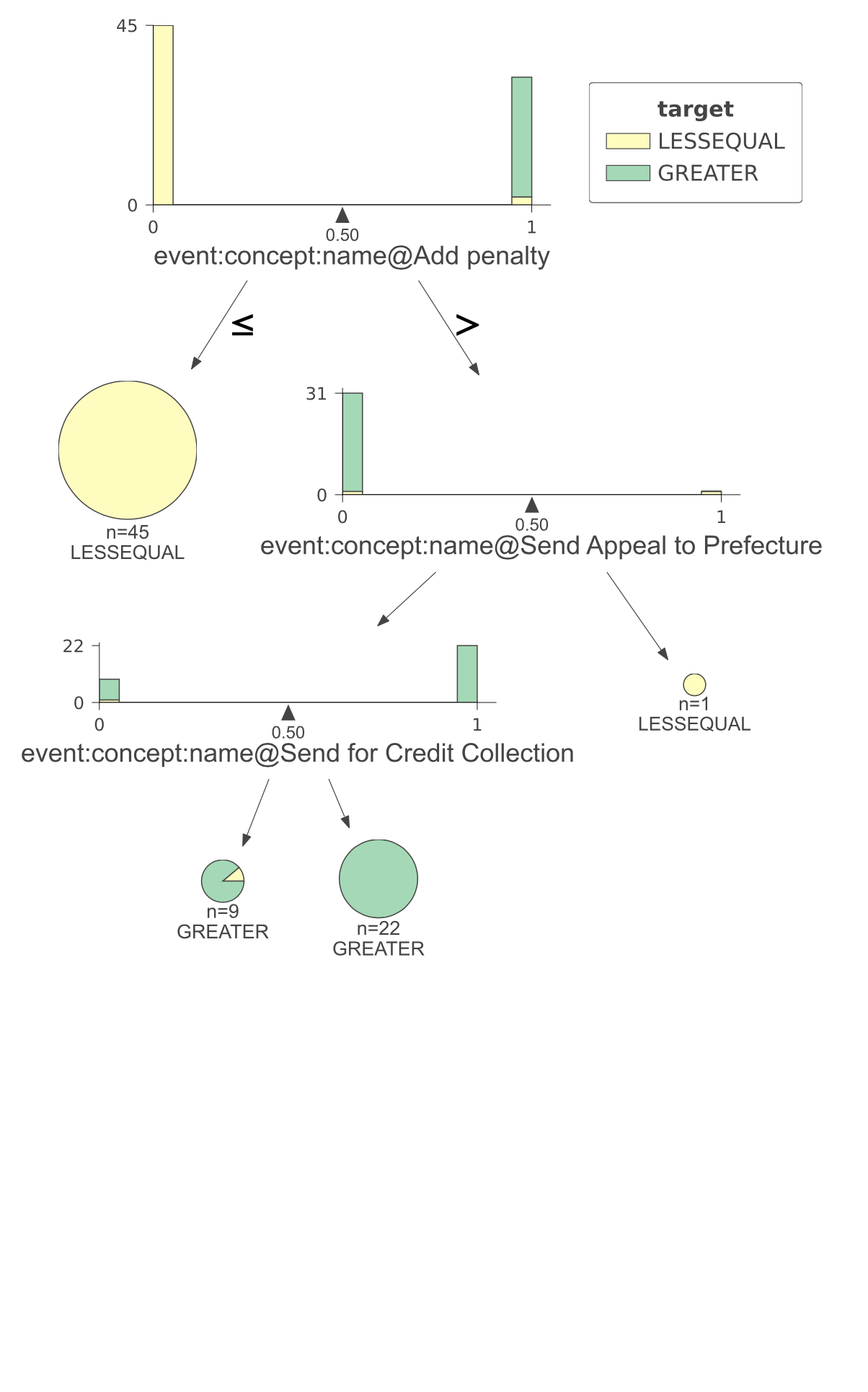}
  \caption{Balanced event log}
  \label{fig:DTsfig2}
\end{subfigure}
 \caption{Two \textit{decision trees} generated from our sample event log. In \ref{fig:DTsfig1} the data in input conforms to the case distribution observed in the event log. As a consequence, the most frequent variants take the lion's share and the numeric feature \textit{amount} decides multiple split points. In \ref{fig:DTsfig2} data is balanced oversampling those variants with low occurrence. The split points in the tree use categorical features only. Decision tree is an example of an algorithm significantly affected by uncritical training  using the case distribution of event logs. }
 \label{fig:DT}
\end{figure}
 
\section{Illustrative Examples}\label{illustrative}
 
We will now use two examples to illustrate the concepts introduced above.

The first example refers to the real-live event log of road traffic fines~\cite{deroad}. 
The events captured in the event log include creating a fine notice, recording the penalty amount, verifying if the payment is received, registering an appeal to the prefecture, and others. The reader interested in more details is referred to \cite{10.1007/978-3-319-39696-5_23}. As illustrated in Table~\ref{tab:1}, the occurrence of trace variants follows a Pareto distribution with only $4$ variants covering more than 88\% of the recorded cases and with $100$ variants that have a single occurrence. The most occurring variant is $\langle Create~Fine,~Send~Fine,~Insert~Fine~Notification,~Add~Penalty,$ $~Send~for~Credit~Collection \rangle^{56482}$, the second is $\langle Create~Fine,~Payment \rangle^{46371}$, the third is $\langle Create~Fine,~Send~Fine \rangle^{20385}$, and so on. 

Let us now try to develop predictive analytics on this event log. For example, we could ask ourselves why certain cases exhibit a duration that is significantly longer than others. To study the problem, we are interested in searching for patterns correlated to long duration. Using encoding, we can represent the cases in the event log as vectors composed of categorical data, such as the executed activities, and of numerical data such as the number of penalties and the trace duration\footnote{The methods used for encoding the event log in a vector space are available in the PM4PY library \url{https://pm4py.fit.fraunhofer.de/documentation\#decision-trees}}. A decision tree can then be used to highlight the factors influencing case duration. We express it as a simple binary problem: being below or above a threshold of 200 days. Figure \ref{fig:DT} illustrates the results we obtain. Figure \ref{fig:DTsfig1} presents a decision tree conforming to the case distribution observed in the event log. 
The entire set of cases in $L$ is encoded in $\mathcal{X}$.
As a consequence, the most frequent variants take the lion's share of the examples used to train the decision tree. Figure \ref{fig:DTsfig2} presents the decision tree obtained by balancing the case distribution among variants, oversampling those variants with low occurrence.
This is, for example, achieved by creating $\mathcal{X}$ taking an equal number of occurrences to the traces in $L$.

Because the split points of the tree are chosen to best separate examples into two groups with minimum mixing, the cases with low occurrence tend to be ignored. 
Indeed, the tree in Figure \ref{fig:DTsfig1} relies on the numeric feature \textit{amount} to decide on multiple split points. On the contrary, the tree in Figure \ref{fig:DTsfig2} defines the split points using categorical features only. This is due to the fact that the variants not associated with a penalty amount were quite rare, and by increasing their representation for balancing the data set we prevented the algorithm to use the penalty amount as a discrimination feature. 

\begin{figure}
\centering
\begin{subfigure}[b]{0.95\textwidth}
   \includegraphics[width=1\linewidth]{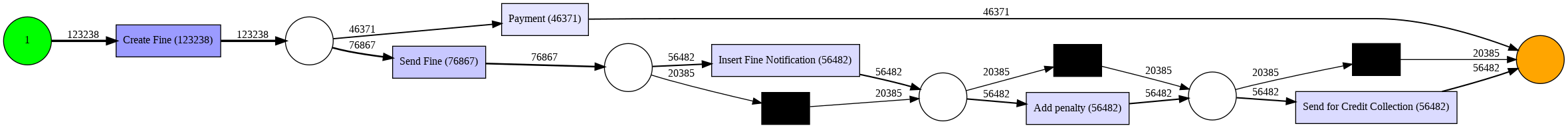}
   \caption{}
   \label{fig:Ng1} 
\end{subfigure}

\begin{subfigure}[b]{0.85\textwidth}
   \includegraphics[width=1\linewidth]{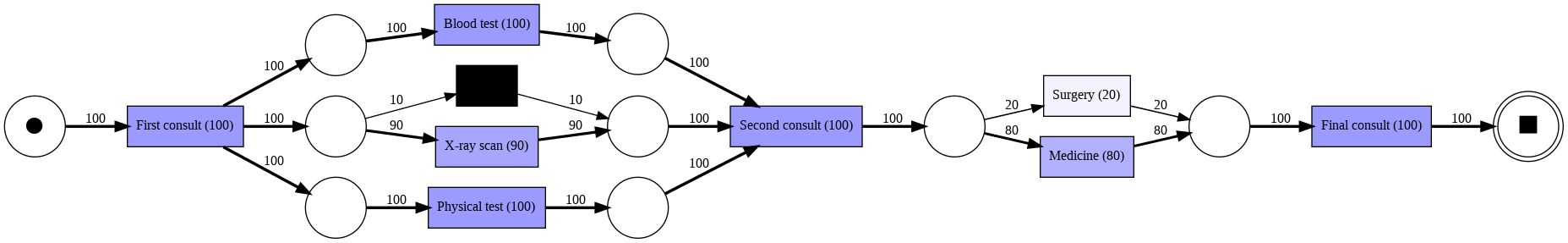}
   \caption{}
   \label{fig:Ng2}
\end{subfigure}

\caption[Alternative routing and parallel execution]{(a) The Heuristic Miner Algorithm~\cite{weijters2006process} was used to discover a model from the ``Road Traffic Fines''~\cite{deroad} event log. The discovered model specifies alternative routes that can be followed to complete the process. In particular, executing {\tt Payment} or {\tt Send Fine} implies the following alternative paths. (b) The Heuristic Miner Algorithm~\cite{weijters2006process} is used to discover a model from the ``Artificial Patient Treatment''~\cite{PMHT20} event log. The discovered model specifies that {\tt Blood test},  {\tt X-ray scan}, and {\tt Physical test} are executed in parallel. Any order can be followed in executing these activities.}
\end{figure}
It is important to note that, in general, we cannot say if proactive balancing is better than using data as they are, and even which is the correct balancing factor to be applied. The strategy to be preferred strongly depends on our goal. If we want to analyse an event log in order to identify procedures that can be automated and learn the decision rule to be used, our interest is in the frequent behaviour. The real distribution of the event log, or even a distribution pruned from rare examples~\cite{van2020pareto}, must address the learning procedure we adopt. If our goal is anomaly detection~\cite{junior2020anomaly} or root cause analysis~\cite{qafari2020root} rare examples have to be represented.

Our next example is related to the need of capturing concurrency (Section~\ref{landscape}). While cases included in an event log are described as sequences of activities, the behaviour they describe should be interpreted differently based on the model that generated them.  To capture control-flow behaviour, one needs to encode the dependency relationships in event logs. 
By executing the Heuristic Miner algorithm~\cite{weijters2006process} on the ``Road Traffic Fines''~\cite{deroad} event log, we observe alternative paths can be followed to complete the process.  If a case includes the execution of the {\tt Payment} activity, it will not include {\tt Send Fine} and the following activities. The same algorithm applied to the ``Artificial Patient Treatment''~\cite{PMHT20} will reveal the concurrent execution of the {\tt Blood test}, {\tt X-ray scan}, and {\tt Physical test} activities. All these activities are required to complete the diagnostic stage, except for {\tt X-ray scan}, which may be skipped, but the order of execution is not relevant. Thanks to process models, PM techniques do consider concurrency. Two sequences $\langle a,b,c  \rangle$ and $\langle a,c,b  \rangle$ can have the same conformance to the model if the model describes $b$ and $c$ as concurrent activities, while the conformance value will be different if $b$ and $c$ are in sequence or relate to alternative paths. Unfortunately, most ML models view event logs merely as sequential data. When cases get encoded into a vector space, the inference the ML model can produce is based on the distance in this space. The distance between $\langle a,b,c  \rangle$, and $\langle a,c,b  \rangle$ is accounted in the same way in the vector space, and we cannot differentiate between the sequences based on the reference process model. This limitation impairs capturing concurrent behaviour that is not detected by simply matching the two sequences. In terms of our example, an ML procedure could effectively predict the lead time of a case knowing that the {\tt Payment} activity was executed. Training an ML algorithm to predict the conformance to the diagnostic protocol of a delivered treatment is more complex, and will require a higher amount of training data, as the ML model needs to incorporate examples on the equivalence of the different orders of execution of the {\tt Blood test}, {\tt X-ray scan}, and {\tt Physical test} activities. Encoding this equivalence in vector space spaces, for example, defining suitable pictograms to feed a CNN, is still an open challenge.

\section{Toward an Integrated Methodology}\label{method}
Guided by the above considerations about encoding, we will now outline the strategy to be used to properly integrate PM and ML. In the previous sections, we argued that when PM tasks are mapped to ML tasks, PM-specific assumptions should drive the construction of training functions and hyper-parameters selection. 
Simple ML classification and regression algorithms model the data by a single Gaussian grounded on mean and co-variance. On the other hand, kernel methods like Gaussian Processes and Support Vector Machines, have opened the possibility of flexible models that are practical to work with, but require non-trivial hyper-parameter tuning to fit behavioural data\cite{melkumyan2011multi}.\\

\begin{figure}[ht!]
    \centering
    \includegraphics[width=1\textwidth]{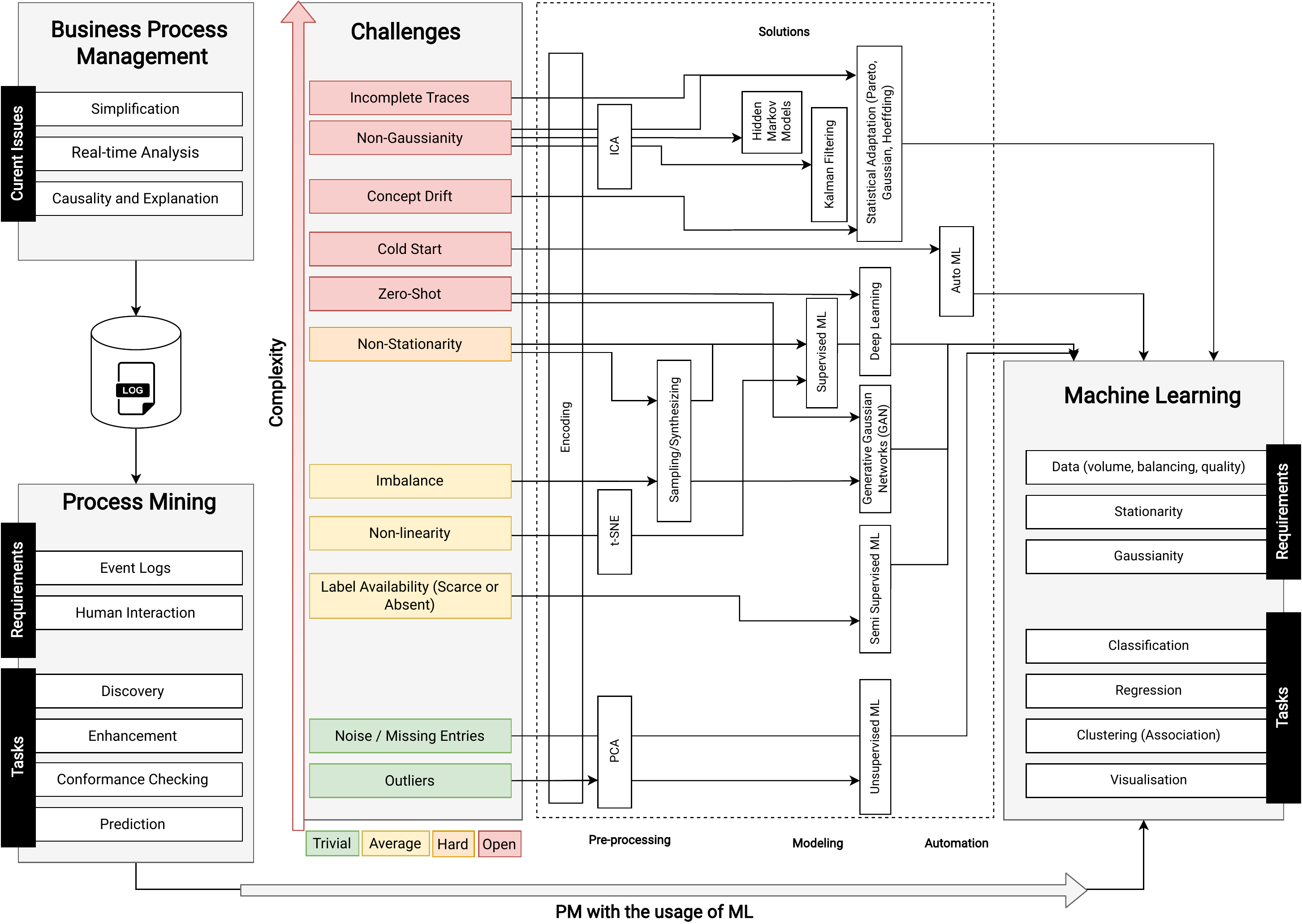}
    \caption{From task to Task, an overview of PM and ML relationship}
    \label{fig:overview}
\end{figure}

Figure \ref{fig:overview} provides a synoptic view of mapping PM tasks to ML ones.

As an example of non-trivial mapping, let us consider the non-linear relationship between data samples and the expected outcomes addressed by robust ML algorithms with adjusted hyper-parameters. At this point, linear projections as PCA are not effective as t-SNE visualisation \cite{van2008visualizing} to obtain insights from the data. 
Other challenges with moderate difficulty are related to label availability and imbalanced scenarios \cite{junior2020anomaly}. In this case, semi-supervised ML techniques and generative models can tackle the label issue, as well as sampling or synthesising methods are the second ones. Problems related to data quality, in which the difficulty is to build an approximation to have a proper data distribution accentuated, can be solved by enlarging the training data and by a proper tuning of the ML algorithm. Alternatively, the training process can be enriched using generative models~\cite{theis2020adversarial}. To handle the difficulties outlined in Section~\ref{intro}, when using non-pictorial traces representation ``process-friendly'' GANs can be considered, like Sequence GANs (SGANs), in which the adversarial samples are designed from discrete sequences, like events. The application of GANs is not limited to data augmentation, as it can be used also for improving data quality for process model generalisation~\cite{theis2020adversarial}. Preliminary results are available on using GAN-generated data to improve predictive tasks (e.g., lead time of incomplete cases) under an adversarial  framework \cite{taymouri2020predictive}.

Coming from non-stationary process behaviour, sampling methods are a promising way to reduce the impact of non-stationary distributions of event log data \cite{cheung2019learning}. 
After bringing the  data to at least a near-stationary behaviour, the business process can naturally change its pattern over time, leading to a burdensome problem called concept drift \cite{bose2011handling,carmona2012online,baier2020handling,ceravolo2020evaluation}. In dealing with this problem, a significant part of the PM community has focused on detecting and managing its onset. Regardless of the success of these attempts, we still consider this problem an open issue, since the event data stream is modelled as a complete trace stream, known from the start to the end activity. In reality, the drift onset occurs at an arbitrary position of the event stream, well before the endpoint is reached and the rest of the trace is known. Some researchers are addressing this information deficiency by using statistical adaptations based on the Hoeffding Bounds  \cite{domingos2000mining}. In principle, it is possible to rely on statistical assumptions about the confidence interval of the data to make a decision on the drift onset. In other words, it is possible to create ML models and perform predictions supported by an approximated conjecture about the future, obtained from the available event log data.
The use of ``stateful'' ML models with memory, in particular, Deep Learners based on the LSTM architecture, could enable handling drifts. However, this kind of challenge demands experienced ML practitioners and a robust computational structure.

\subsection{Hyper-parameter Tuning}
Once a class of ML models has been chosen, hyper-parameter tuning must be performed to instantiate the ML model that delivers the desired accuracy (and possibly some required non-functional properties, like explainability). Searching the model space by trial and error can be burdensome. Automated Machine Learning (AutoML) is a reasonable alternative to tackle these problems grounded on sharing previous knowledge for similar tasks. AutoML can help to handle the classification problems called Zero-shot~\cite{wang2019survey,ji2019multi} or Cold-start \cite{chemingui2019product}, for which little context information (and even the complete list of classes) may not be available at the start of the training, by taking advantage of meta-features and information on similar models, akin to how human experts start an old-fashioned search for desirable models driven by their experience on related tasks~\cite{hu2019correction}.  Some PM research works based on AutoML discuss how to find a suitable PM pipeline by recommending steps. For example, \cite{tavares2021process} proposed a solution to suggest the encoding method, since the higher number of methods might lead to a tricky selection. Furthermore, there are encoding methods able to fit particular data. It is remarkable that traditional process mining tasks can be leveraged when matched with intelligent decision support approaches.

\subsection{Final Recommendations}
In this final section, we present a set of recommendations that aim to be valuable for both PM practitioners and researchers.

\subsubsection{RECOMMENDATION 1: Choose data representation carefully}
When working with PM data structures, it is crucial to carefully translate them into a metric feature space that can be manipulated by ML algorithms. Additionally, it is important to preserve context information, such as control-flow constraints, which are essential for process analysis. The choice of encoding techniques should align with problem-specific goals and constraints.
\subsubsection{RECOMMENDATION 2: Fit the data distributions} PM often deals with non-Gaussian, non-stationary distributions. To achieve optimal performance in production, it is advisable to estimate the data distribution instead of relying on the best Gaussian mix approximation. Building training sets interactively poses a significant challenge in PM. Leveraging ML approaches such as AutoML and Active Learning can help reduce the manual burden and improve the process.
\subsubsection{RECOMMENDATION 3: Do not assume the availability of a labelled training set} In business process environments, obtaining pre-existing labelled training sets for PM tasks is uncommon. Constructing a training set by correctly sampling the data space is essential, particularly due to the high diversity of process execution conditions in PM tasks.
\subsubsection{RECOMMENDATION 4: Consider zero-shot learning} During the training of ML models, the complete set of possible outcomes (co-domain of $f$) may only be partially known. For instance, in process optimisation, the cost of certain sequences may not be available at the time of training the regression model. It is essential to assess the completeness of the available information when formulating the problem statement to ensure the quality of model inference.
\subsubsection{RECOMMENDATION 5: Ensure minimum ML quality at an early stage via constraints} As the estimation of data distribution converges over time, an extended convergence period is unacceptable as it results in a high model error during training. It is possible to impose control flow constraints on ML models when they are known in advance based on domain requirements and regulations.
\subsubsection{RECOMMENDATION 6: Incorporate domain knowledge} Domain knowledge plays a critical role in effective PM. Integrating domain-specific information and constraints into ML models can significantly enhance their performance and interpretability. It is important to actively involve domain experts in the feature engineering and model validation processes.
\subsubsection{RECOMMENDATION 7: Evaluate model interpretability} PM tasks often require interpretable models to gain insights into process behaviour and make informed decisions. It is essential to evaluate the interpretability of ML models and choose algorithms that provide transparent explanations of their predictions. This becomes particularly crucial when dealing with critical processes or compliance and regulatory requirements.
\subsubsection{RECOMMENDATION 8: Continuously monitor and update ML models} Process environments are dynamic, and changes over time can impact the performance of ML models. Establishing a framework for monitoring and evaluation allows the assessment of models' performance and facilitates their timely updates as needed. Continuous learning and retraining of models ensure their accuracy and relevance in evolving process scenarios.
\subsubsection{RECOMMENDATION 9: Share knowledge and best practices} Promote knowledge sharing and collaboration within the PM community. Encourage the dissemination of successful case studies, research findings, and best practices to foster learning and advancement in the field. Engage in conferences, workshops, and online forums to connect with fellow practitioners and researchers and stay updated with the latest developments in PM. 

By following these recommendations, PM practitioners and researchers can improve the effectiveness and efficiency of process mining applications, enabling better process understanding, optimisation, and decision-making.

\section{Conclusions}\label{conc} 

The growing use of ML methods in PM necessitates a robust and comprehensive methodology for integrating these algorithmic techniques. This paper aimed to address the challenges associated with the ML/PM mapping and identify the fundamental principles for establishing a methodological foundation in this field.
Through the analysis conducted in this study, we have provided a set of recommendations that can guide practitioners and researchers in effectively applying ML to PM tasks. These recommendations encompass various aspects of the PM process, from data representation to model evaluation and monitoring.
By following these recommendations, PM practitioners and researchers can enhance the effectiveness and efficiency of their ML-driven process mining applications.
It is important to acknowledge that the field of ML in PM is constantly evolving, and new challenges and opportunities will continue to arise. As such, ongoing research and collaboration among practitioners and researchers are crucial to refine and expand upon the proposed recommendations.
By embracing a methodological foundation that integrates ML techniques in PM, we can unlock the full potential of process mining and leverage the power of data-driven insights to drive process understanding, optimisation, and decision-making in various domains and industries.

\bibliographystyle{unsrt}
\bibliography{references}

\end{document}